\documentclass[12pt]{article}

\usepackage{multirow}

\usepackage{mathptmx}
\usepackage{graphicx,cite}
\usepackage{times}
\graphicspath{{./imgs/}}
\usepackage{hyperref}
\usepackage{amsmath}
\usepackage{amssymb}
\usepackage{booktabs}
\usepackage{url}
\usepackage{breakurl}
\usepackage{subfigure}
\usepackage{hyperref}
\usepackage{gensymb}
\usepackage[final]{microtype}
\usepackage[T1]{fontenc}
\usepackage{soul}
\usepackage{color}

\usepackage{times}

\newenvironment{sciabstract}{%
\begin{quote} \bf}
{\end{quote}}

\newcounter{lastnote}

\title{Comparative Benchmarking of Causal Discovery Techniques}

\author
{Karamjit Singh, Garima Gupta, Vartika Tewari and Gautam Shroff\\
\\
{TCS Research, New Delhi, India}\\
\\
{\{karamjit.singh, gupta.garima1, vartika.tewari, gautam.shroff\}}@tcs.com
}

\date{}

\begin{document} 


\baselineskip24pt


\maketitle 

\section*{Abstract}
\begin{sciabstract}
%\abstractname{}
 In this paper we present a comprehensive view of prominent causal discovery algorithms, categorized into two main categories (1) assuming acyclic and no latent variables, and (2) allowing both cycles and latent variables, along with experimental results comparing
them from three perspectives:
(a) structural accuracy, (b) standard predictive accuracy, and (c) accuracy of counterfactual inference. For (b) and (c) we train causal
Bayesian networks with structures as predicted by each causal discovery technique to carry out counterfactual or standard predictive inference.
We compare causal algorithms on two publicly available and one simulated datasets having different sample sizes: small, medium and large. Experiments show that structural accuracy of a technique does not necessarily correlate with higher accuracy of inferencing tasks. Further, surveyed structure learning algorithms do not perform well in terms of structural accuracy in case of datasets having large number of variables.
\end{sciabstract}

% In setting up this template for *Science* papers, we've used both
% the \section* command and the \paragraph* command for topical
% divisions.  Which you use will of course depend on the type of paper
% you're writing.  Review Articles tend to have displayed headings, for
% which \section* is more appropriate; Research Articles, when they have
% formal topical divisions at all, tend to signal them with bold text
% that runs into the paragraph, for which \paragraph* is the right
% choice.  Either way, use the asterisk (*) modifier, as shown, to

\section{Introduction}
% !TEX root = ../template.tex
Given a set of random variables $V$, the problem of identifying \textit{causal networks}~\cite{pearl2003causality}\cite{spirtes2000causation}, with each variable as a node and each edge representing a cause and effect relationship between variables is called \textit{causal discovery}. Causal discovery is of paramount interest in many 
applications domains: For example, in medical diagnosis researchers are concerned with discovering the conditions, events, or genes that are likely to cause a certain disease. Similarly, in the biological domain one is interested in finding cause and effect between the expression levels of pairs of genes, proteins, or gene-protein pairs.
In marketing, on the other hand, causality has been used for evaluating the efficacy of marketing campaigns as well as evaluate alternative strategies 
without actually trying them out in practice, via `counterfactual' inference.

Once causal relationships have been learned it is possible to derive causal models that allow us to
infer or predict the variables of interest for many practical scenarios: For example, once the causes
of a disease are discovered, these can be further used to predict its occurrence at earlier stages than otherwise. Another important use case of causal model learning and inferencing is in \textit{optimal product mix} problems: Consider a case where dealer makes quarterly decision of how many of each product to purchase in order to maximize the profit for the quarter. In such cases quantity of each product to be purchased is the decision variable which alongwith other factors can be modelled as a causal model to infer the effect, i.e, quantity sold for each product. Further a linear mix of these causal predictions can be used to maximize profit and obtain optimal product purchase mix subject to certain constraints.

Another advantage of causal models are 
for what-if analysis , or counterfactual inference \cite{pearl2003causality}. For example consider an online advertisement setting: 
answering whether a proposed modification in the deployed ad-placement algorithm will result in better results.
A change in the algorithm can be seen as an `intervention' in the language of causality \cite{pearl2003causality}: The simplest way to test 
such a modification is to deploy it in the field and run an A/B test; in most cases this involves time and expense, both for
implementation as well as possible business losses if a modification is unsuccessful. An alternate approach would be to pose this as a counterfactual
question that could be answered using previously collected data without actually deploying the change; causal models enable such analysis.

\subsection{Overview of Causal Discovery Techniques}

In general causal discovery requires controlled experiments via interventions, which in many cases is too expensive, unethical, or technically impossible to perform. For example, it would be considered unethical to force a random selection of people to smoke in order to establish smoking as a cause for lung cancer. The development of methods to identify causal relationships from purely observational data is therefore desirable. Many algorithms have been developed using observational data to infer causal relations, particularly in the area of graphical causal modeling \cite{koller2009probabilistic}~\cite{pearl2003causality}. However these algorithms can identify structures only
upto Markov equivalence class \footnote{For example the three models $X \rightarrow Y \rightarrow Z$, $X \leftarrow Y \leftarrow Z$, and $X\leftarrow Y\rightarrow Z$ all encode the same conditional independence, $X\bot Z\mid Y$ and are Markov equivalent.}. 

Data generated from controlled experiments, or interventions further allow us to distinguish between causal graphs that are Markov equivalent. However, in certain situations when number of experiments that can be performed are limited, finding an optimal set of experiments that are most informative towards uncovering the underlying structure is very crucial. This class of problem is called \textit{active learning}~\cite{tong2001active}. The intuition behind active learning is that every variable is not equally informative when intervened. For instance, if $X$ does not have any children in every graph of a Markov equivalence class, perturbing $X$ will not lead to any visible impact that can further distinguish the graphs.

Causal Bayesian networks~\cite{pearl2003causality}, a commonly used form of causal model, are often restricted to be acyclic. However, most biological networks have cycles or feedback loops in them. Therefore, another class of techniques that allow for cycles and feedback loops are very important especially in biological domain. Many algorithms in causal discovery have another important assumption in that latent variables are ignored: It is widely recognised that incorporating latent or hidden variables is a crucial aspect of causal modelling that can provide a more succinct representation of the observed data.%; in causal language, possibly many observed variables can be seen as emanating from fewer albeit hidden, causes. 

\subsection{Key Contributions and Outline}
In this paper we do the following:
\begin{enumerate}
 \item Provide an overview of  causal discovery techniques, classified into two main categories: 1) assuming acyclicity and no latent variables, and 2) 
 allowing for both cycles and latent variables. In the first category we further divide techniques into three classes, 1) \textit{observations only}: algorithms which use observation data only to infer causal structure, 2) \textit{use mix of observation and intervention}: algorithms which use both observations and intervention data, and 3) \textit{active learning}.
 \item  Present a comparative experimental benchmarking of causal discovery techniques using performance metrics constructed from three perspectives: 1) Accuracy
 of structure learning, 2) Predictive accuracy, and  3) Counterfactual accuracy. Our experiments have been performed on three datasets (two real and one simulated) having different sample sizes; large, medium, and small.
\end{enumerate}

After stating the background definitions in Section~\ref{sec:back}, we briefly discuss the prominent algorithms of both categories and their advances in Section~\ref{sec:estimating_structures},~\ref{sec:CylesAndHidden}. In Section~\ref{sec:metrics}, we define the three types of performance metrics which are used for comparison. 
In Section~\ref{sec:experiments} we present benchmark results using the prominent structure learning techniques from each category over three datasets of different sample sizes. We also benchmark both predictive accuracy and structural accuracy on all three datasets and counterfactual accuracy on a simulated dataset, in each case
using Bayesian network models having structure as predicted by each causal discovery technique. Finally, after discussing the related work in section~\ref{sec:rel}, we conclude in section~\ref{sec:conc} by discussing the merits and demerits of the techniques surveyed.
\label{sec:intro}
\section{Background}\label{sec:back}
% !TEX root = ../template.tex
We consider a graph with nodes having two types of edges: directed edges $(\rightarrow)$ and undirected edges $(-)$. 
%There can be atmost one edge between two nodes.
A graph having all directed (undirected) edges is called directed (undirected) graph. A directed graph with no cycles is called \textit{directed acyclic graph (DAG)}. A graph with mix of directed and undirected edges is called \textit{partial directed graph} and along with the acyclic assumption is called \textit{partial directed acyclic graph} (PDAG). The \textit{skeleton} of a partial directed graph is the undirected graph that results from replacing all directed edges by undirected edges. 

%Two nodes in a graph are \textit{adjacent} if they are connected by an edge. If $X \rightarrow Y$, then $X$ is a parent of $Y$, and $Y$ is a child of $X$. The adjacency set and parents set of a node $X$ is denoted by $adj(X)$ and $par(X)$, respectively. A \textit{path} is a sequence of distinct and adjacent nodes. A path is directed, if all edges on the path are directed in the same direction. The \textit{ancestors} of X are all nodes from which a directed path leads to X. Correspondingly, the \textit{descendants} of X are all nodes to which a directed path leads from $X$. 
%The path $X$ $-$ $Y$ $-$ $Z$, where $X$ and $Z$ are not adjacent is called an unshielded triple. 

A causal Bayesian network (CBN) is a Bayesian network with each edge representing cause and effect relationship from parent to child. A CBN satisfies \textit{Causal Markov property}: Every node is independent from its non-descendants given its parents. 

The node Y on the path $X \rightarrow Y \leftarrow Z$ is called \textit{collider}. If $X$ and $Z$ are not adjacent, $Y$ is called an unshielded collider. In CBN, two nodes $X$ and $Y$ are said to be \textit{d-separated} by a set S, if on every path X and Y (1) atleast one non-collider is in S, or (2) atleast one collider is not in S nor has a descendant in S. A causal graph $G$ with probability distribution $P$ satisfies \textit{faithfullness condition} if and only if every conditional independence relation true in $P$ is entailed by the Causal Markov Condition applied to $G$. Two DAGs that have the same d-separation properties are said to be Markov equivalent, and class of such DAGs is called \textit{Markov equivalence class}. A Markov equivalence class can be described uniquely by a \textit{completed partially directed acyclic graph (CPDAG)}. 
%The skeleton of the CPDAG is defined as follows: Two nodes $X$ and $Y$ are adjacent in the CPDAG if and only if, in any DAG in the Markov equivalence class, $X$ and $Y$ cannot be d-separated by any set of the remaining nodes. 
The orientation of the edges in the CPDAG is as follows: A directed edge $X \rightarrow Y$ in the CPDAG means that the edge $X \rightarrow Y$ occurs in all DAGs in the Markov equivalence class. An undirected edge $X - Y$ in the CPDAG means that there is a DAG in the Markov equivalence class with $X\rightarrow Y$, as well as with $X\leftarrow Y$.

Ancestral graphical models are used when causal structure allows for latent variables. These latent variables can be i) \textit{hidden confounder:} variables which are a common cause of at least two observed variables, ii) \textit{hidden selection variables}: hidden variables that are caused by at least two observed variables. A \textit{maximal ancestral graph (MAGs)} consists of nodes representing (observed) variables and edges, which might be directed, undirected or bi-directed. Thus, each
end of an edge can have the edgemark \textquoteleft tail \textquoteright or \textquoteleft head \textquoteright. If on the edge between X and Y there is a head mark at X, then X is not a ancestor of Y (or any selection variable) in the underlying DAG. If on the other hand there is a tail mark at X, then X is an ancestor of Y (or any selection variable). MAGs encode conditional independence information using m-separation, a graphical criterion which is very closely related to d-separation for DAGs. As with DAGs, we usually cannot identify a single MAG given observational data, rather several MAGs which encode the same set of conditional independence statements form a Markov equivalence class represented by a \textit{partial ancestral graph (PAG)}.% A PAG is like a MAG with the only difference that some edge marks are unknown (and usually represented by circles).
% The most frequently used causal models belong to two broad families: (1) causal Bayesian networks \cite{cooper1999overview}\cite{spirtes2000causation}, and (2) structural equation models (functional causal models) \cite{duncan2014introduction}. In this paper, we focus on reviewing the algorithms for casual discovery using causal Bayesian networks.
\section{Acyclicity and no Latent Variables}\label{sec:estimating_structures}
Many algorithms have been proposed for learning CBNs with the assumptions of acyclicity and no latent variable. In this section, we discuss the three classes of algorithms under these assumptions.

\subsection{Using observations only}
Causal discovery from observation data has been widely accepted as the best alternative to randomized controlled experiments, since observational data can often be collected cheaply and is easily available. The algorithms for causal discovery can be grouped into three broad categories: constraint-based, score-based algorithms and hybrid methods.

\textbf{Constraint-based methods:} Constraint-based algorithms learn causal Bayesian networks with conditional independence tests through analyzing the probabilistic relations entailed by the Markov property of networks~\cite{pearl2003causality}. Given an observational data set D defined on a variable set V , the constraint-based algorithms consists of three key steps: (1) uncovering the entire skeleton, (2) discovering v-structures, and (3) orienting edges as many as possible. Mainly constraint-based algorithms assume following three conditions: 1) causal sufficiency, 2) faithfulness and Markov condition, and 3) reliable independence tests.
The independence tests generally can be implemented using G2 test \cite{agresti2011categorical}, mutual information \cite{wyner1978definition}, and Fisher$'$s Z-test~\cite{pena2008learning}. %The the G2 test and muual information are both for dealing with discrete (categorical) data, while the Fisher’s Z-test handles continuous (numeric) data.

\textbf{PC}~\cite{spirtes2000causation} is one of the most popular constraint-based algorithm for learning causal graph. It starts with a complete undirected graph i.e., every node is connected to every other node. Then, for each edge (say, between $X$ and $Y$ ) it tests, whether there is any conditioning set S, so that X and Y are conditional independent given S, denoted by $X \perp Y |S$. If such a set 
%(called conditioning set) 
is found, the edge between $X$ and $Y$ is deleted. The algorithm considers conditioning set of increasing size, starting from empty, until the size of the conditioning set is larger than the size of the adjacency sets of the nodes. % ($X$ and $Y$). 
This step gives the skeleton structure. %when using perfect conditional independence information (the tests are reliable and the data instances are enough large). 
Further, after applying certain edge orientation rules, the output of the PC algorithm is the estimated CPDAG.% This output depends on the ordering of the variables (except in the limit of an infinite sample size), since ordering determines which conditional independence tests are done. %The complexity of the algorithm for a DAG G is bounded by the largest degree over all pairs of vertices in G .

%PC algorithm is closest to SGS~\cite{spirtes2000causation} algorithm in terms of approach with the difference is at the edge elimination step: SGS tests every possible conditioning set at every order of conditioning, while the PC tests only conditioning sets involving variables that are connected by direct or indirect paths to the variable under test, making PC algorithm computationally more efficient.

Many variants have been developed which improve upon the PC algorithm. Conservative PC algorithm (CPC) \cite{ramsey2012adjacency} improves the robustness in orientation phase and captures violations of the Orientation-Faithfulness assumption.Adjacency Conservative PC algorithm (ACPC )~\cite{lemeire2012conservative} relaxes the Adjacency-Faithfulness assumption and is experimentally shown to be superior to PC and CPC when pseudo-independent relations and equivalent edges are violated. PC-stable algorithm~\cite{colombo2012learning} yields order-independent skeletons but requires more independence tests and therefore results in even longer running time than the PC algorithm. %Worst case running time of PC algorithm is exponential to the number of variables, and thus is computationally inefficient in case of high dimensional data.  Parallelized versions of PC algorithm and PC-stable algorithm \cite{le2016fast}~\cite{le2015parallelpc} using parallel computing technique produce same outputs as original algorithms, but are much more time efficient.

\textbf{Score-based methods:} Score-based algorithms assign a score to each candidate network for measuring how well the candidate network fits dataset \cite{chickering2002learning}~\cite{cooper1992bayesian}. %Score-based methods learn the CPDAG by greedily searching for an optimally scoring DAG. 
Greedy equivalence search (GES) algorithm~\cite{chickering2002optimal} is one of the prominent example of score-based algorithms. GES has two phases: forward and backward phase. The forward phase starts with an empty graph, and  sequentially adds single edges, each time choosing the edge addition that yields maximum improvement of the score, until the score can no longer be improved. The backward phase starts with the output of the forward phase, and sequentially deletes single edges, each time choosing the edge deletion that yields maximum improvement of the score, until the score can no longer be improved. %GES is computationally more efficient than other traditional score-based methods as it searches over the space of all possible CPDAGs, instead of over the space of all possible DAGs. 
GES assumes two properties: 1) score equivalent, and 2) decomposability. Score equivalent property makes sure that every DAG in a Markov equivalence class gets the same score and decomposability property of scoring criterion allows fast updates of scores during the forward and the backward phase, e.g., regularized log-likelihood such as BIC~\cite{schwarz1978estimating}. Finally, consistency property of scoring criterion ensures that the true CPDAG gets the highest score with probability approaching one.

GES can be improved by including a turning phase of edges as well~\cite{hauser2012characterization}. An analysis of the penalized maximum likelihood estimator for sparse DAGs in high dimensions which also avoids the faithfulness assumption is presented in \cite{van2013ell}. A parallelized version of GES for continuous variable called FGS is presented in \cite{ramsey2015scaling}

\textbf{Hybrid methods:} Apart from constraint-based and score-based algorithms, there is another class of algorithms called hybrid methods. Hybrid methods learn the CPDAG by combining constraint-based and score-based methods. Typically, they first estimate (a supergraph of) the skeleton of the CPDAG using conditional independence tests, and then apply a search and score technique while restricting the set of allowed edges to the estimated skeleton. A prominent example is the Max-Min Hill-Climbing (MMHC) algorithm~\cite{tsamardinos2006max}. %The idea of MMHC is to find the skeleton based on a constraint-based search and then use a search-and-score approach to orient the edges. 
In the empirical study, authors showed MMHC outperform PC and GES with structural hamming distance (SHD) as a measure. 

%For small dimensional datasets,~\cite{perrier2008finding} learns an optimal Bayesian network (BN) (i.e., it converges to the true model in the sample limit).~\cite{kojima2010optimal} proposed to perform an optimal search on several clusters in order to scale up to larger networks. Despite interesting improvements in terms of score and SHD on several benchmark BNs, the reported running time is about $10^3$ times longer than MMHC on an average. \cite{gasse2012experimental} H2PC and MMHC share exactly the same search and score (SS) procedure to allow for fair comparisons. While MMHC is based on Max-Min Parents and Children (MMPC) to learn the parents and children of a variable, H2PC is based on a subroutine called Hybrid Parents and Children (HPC), that combines ideas from incremental and divide-and-conquer constraint based methods. In their empirical study, they claimed that H2PC outperforms MMHC in reconstructing the original DAG with little overhead in terms of of running time over MMHC. 

% Hybrid methods have a huge computational advantage over score based methods due to the constraints on search space. However, this comes at the cost of inconsistency or at least at the cost of a lack of consistency in proofs. ARGES \cite{nandy2015understanding} proposed consisent hybrid modification of GES where the search space depends mainly on the estimated conditional graph. But the search space changes adaptively depending on the current state of the algorithm.

\subsection{Using observation and intervention data}\label{sec:observation and interventiondata}
%There are two major types of interventions, 1) \textit{Perfect intervention}: it occurs when an intervention sets a variable to a fixed value, and 2) \textit{Imperfect intervention}: the variable does not take a fixed value, but takes a different distribution of values than the original one, upon intervention.
The concept of Markov equivalence class must be extended to capture DAGs which are equivalent given observational data and data from specific interventions. However combining observations and intervention data is a challenging task due to the fact that intervention data samples from distribution which is not identically distributed as the observational distribution.

In~\cite{hauser2015jointly}, authors point out that even when intervention data is given, the causal structure is generally not fully identifiable. However, the intervention data helps in identifying at least some parts of the causal structure, while other parts of the structure might remain ambiguous or could be uniquely identified if results on different
intervention were available. %Therefore, the concept of Markov equivalence class must be extended to capture DAGs which are equivalent given observational data and data from specific interventions. %The authors provide this extension in the form of the interventional essential graph, generalizing the CPDAG. 

Greedy interventional equivalence search (GIES)~\cite{hauser2012characterization} is a prominent example of causal Bayesian structure learning algorithm which uses both observation and intervention data. GIES extension of score based method GES. It is based on optimization of BIC-score greedily moving through the space of essential interventional graphs with repeated forward, backward and turning phases.

\subsection{Active learning}\label{sec:active learning}
Active learning in the Bayesian setting was introduced by Tong and Koller in \cite{tong2001active}. %(we refer it as TK algorithm), 
The algorithm proposed decision-theoretic frameworks based on the expected reduction in uncertainty over edge directions. It builds upon \cite{friedman2000being}, which does MCMC over total orderings of nodes, instead of over DAGs, which makes the complexity of size $O(n!)$ (size of the space of ordering) instead of $O(2^{n^2})$ (size of the space of DAGs). \cite{murphy2001active} 
%(we refer it as Murphy algorithm) 
is an another popular work in active learning space which is also a decision-theoretic framework but on the expected change in posterior distribution over graph. Difference between both the algorithms lies in the form loss function, where \cite{murphy2001active} uses more simplistic loss function having no assumption about the form of it. \cite{tong2001active}s algorithm is faster than \cite{murphy2001active} because the former uses MCMC in the smaller space of orderings.
%TK algorithm is faster than Murphy as it computes $\sum_y$ analytically rather than using importance sampling. And secondly, it uses MCMC in the smaller space of orderings. 
While both of these approaches have been shown to be effective, they have been studied only in the context of discrete Bayesian networks. For continuous variables, Guassian Bayesian networks (GBNs) have been used in various papers \cite{grzegorczyk2010introduction}\cite{maathuis2010predicting}\cite{rau2013joint}, where each variable is continuous and is modeled as a function of its parents with added Gaussian noise. \cite{cho2016reconstructing} proposed an efficient active learning algorithm for biological networks based on the framework of Murphy algorithm. %It is a first Bayesian active learning algorithm for GBNs, where 
The informativeness of each candidate intervention is estimated via Bayesian inference, treating the graph as a latent random variable, and the most informative intervention is chosen. It also introduces an optimization technique unique to GBNs that leads to significant runtime improvement.

%\subsection{Structure learning using observation data}\label{sec:observation_data}
%\input{Sections/observation_data}
% \subsection{Structure learning using observation and intervention data}\label{sec:observation and interventiondata}
% \input{Sections/observation_and_intervention_data}
\section{Allowing cycles and Latent Variables}
\label{sec:CylesAndHidden}
\subsection{Structure learning assuming hidden variables}\label{sec:hidden variables}
While learning causal relationship 
%or predicting the outcome of an intervention 
two major challenges are faced, one is the presence of hidden confounders %: hidden variables which are a common cause of at least two observed variables
and the other is selection bias.%: hidden variables that are caused by at least two observed variables. In this section, we will discuss the approaches that assumes the presence of hidden counfounders and selection variables during structure learning.

The Fast Causal Inference (FCI) algorithm~\cite{spirtes1995causal}\cite{spirtes2000causation} was one of the first algorithms that was able to infer causal relations from conditional independence tests in the large sample limit, even in the presence of latent and selection variables. FCI algorithm can be outlined in five steps. The first and second step of FCI is analogous to the PC algorithm. It starts with the complete undirected graph, an initial skeleton and unshielded colliders are found as in the PC-algorithm. Due to the presence of hidden variables, it is no longer sufficient to consider only subsets of
the neighborhoods of nodes $X$ and $Y$ to decide whether the edge $X-Y$ should be removed. Therefore, the initial skeleton may contain some superfluous edges. These edges are removed in the next step (third) of the algorithm. In the third step, \lq\lq Possible-D-SEP\rq\rq is computed. The edges of the initial skeleton are then tested for conditional independence given subsets of Possible-D-SEP. If conditional independences are found, the conditioning sets are recorded as separation sets and the corresponding edges are removed. Thus, edges of the initial skeleton might be removed and the list of all separation sets might get extended. 
%Since Possible-D-SEP can get very large even in sparse graphs, testing all subsets might be very time consuming. 
In step four, unshielded colliders in the updated skeleton are oriented based on the updated list of separation sets. In step five, further orientation rules are applied in order to avoid contradictions in the resulting PAG.

The run time complexity of FCI algorithm is exponential (even in case of sparse underlying DAG), which is a major drawback of FCI. Anytime FCI~\cite{spirtes2001anytime} combats the problem of exponential run time by allowing a trade-off between computational speed and informativeness by setting an additional tuning parameter. Really Fast Causal Inference (RFCI)~\cite{colombo2012learning} algorithm is another improvement of FCI, and does not compute any Possible-D-SEP sets, and thus does not make tests conditioning on subsets of Possible-D-SEP. This makes RFCI much faster than FCI. 
%However, some modifications of step two (unshielded colliders) and step three (further orientation) avoid some of the erroneous causal conclusions PC could make in the presence of hidden variables.
 The causal interpretation of RFCI is sound and consistent in high-dimensional, sparse settings, but slightly less informative than that of FCI. %In \cite{colombo2012modification}, authors show in simulations that the estimation performances of FCI and RFCI are very similar.
FCI+~\cite{claassen2013learning} is an alternate improvement of FCI, which is sound and complete and has polynomial complexity for underlying sparse DAGs.
\subsection{Structure learning assuming loops}\label{sec:loops}
Algorithms which we have discussed till now have assumption of discovering acyclic (non-cyclic) directed graphs (DAGs) from data. Cyclic Causal Discovery (CCD)~\cite{richardson1996discovery} takes cyclic condition into account and is one of the most well known algorithms that assumes cycles but no latent variables. 
%CCD outputs i) A PAG that represents a set of directed graphs that entail the same set of zero partial correlations for all values of the linear coefficients, ii) features common to those directed graphs (DGs) (such as ancestor relations). 
CCD performs a series of statistical tests of zero partial correlations to construct PAG as output. 
%The set of zero partial correlations that is entailed by a linear SEM with uncorrelated errors depends only on the linear coefficients, and not on the distribution of the error terms. 
%There are a number of limitations of this algorithm. First, the
The set of directed graphs contained in a PAG can be large, and while they all entail the same zero partial correlations they may not entail the same joint distribution or even the same covariances. Hence in some cases, the set represented by the PAG will include cyclic graphs that do not fit the data well, which is a limitation of this algorithm. %Therefore, even assuming that the errors are all Gaussian, it is possible to reduce the size of the set of graph outputs by CCD, although in practice this can be intractable.
\section{Performance Metrics}\label{sec:metrics}
% !TEX root = ../template.tex
We compare the performance of causal discovery algorithms from three perspectives: structural, predictive and counterfactual accuracy. We use following measures:

\subsection{Structure learning measures} 
\begin{figure}[h]
\centering
\includegraphics[width = 100mm]{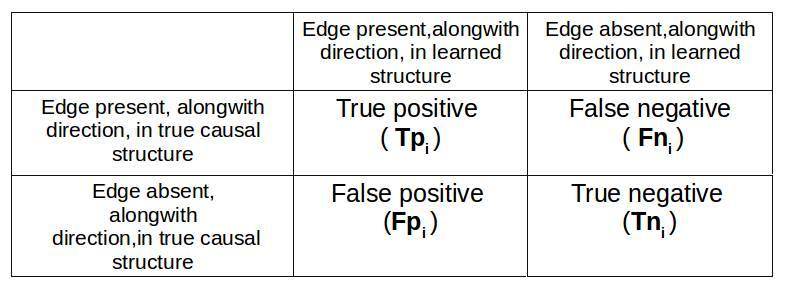}
%\vspace{6pt}
\caption{Confusion matrix, i corresponds to each edge}
\label{fig:confusion_matrix}
%\vspace{-10pt}
\end{figure}

We compare different algorithms based on the learned causal structure compared to the actual causal structure using following standard structure learning measures as follows:

\textbf{F-score:} It is the harmonic mean of precision and recall of the learned structure as compared to original structure. We calculate true positive ($T_{p_i}$), true negative ($T_{n_i}$), false positive ($F_{p_i}$), and false negative ($F_{n_i}$) in the learned structure, based on the directed edges discovered or missed as formally described in Figure~\ref{fig:confusion_matrix}.

\textbf{Area under ROC curve (AUC):} Area under the curve of recall versus false positive rate (FPR) at different thresholds is the desired AUC. 

\textbf{Structural hamming distance (SHD):} The Structural Hamming Distance~\cite{acid2003searching}\cite{tsamardinos2006max} considers two partially directed acyclic graphs (PDAGs) and counts how many edges do not coincide.

\textbf{Structural Intervention distance (SID):}  The SID~\cite{peters2013structural} is based on a graphical criterion only and quantifies the closeness between two DAGs in terms of their corresponding causal inference statements. It is therefore well-suited for evaluating graphs that are used for computing
interventions.

\subsection{Predictive accuracy} Models derived from a causal graph (learned via causal discovery) can be further used for multiple practical applications involving predictions, intervention queries, or counterfactual reasoning. We use normalized root mean square error to measure the predictive accuracy of a Bayesian network
based on a causal graph as follows:% with best \textit{Structural F-score}.

\textbf{Bayesian Network Predictive NRMSE:}  We extend selected PAG/CPDAG to DAG using \cite{dor1992simple}. If no extension is possible, DAG corresponding to the skeleton of the PDAG is taken. Having obtained DAG structure of $m$ nodes, 
%we model the DAG as CBN with $m$ nodes. 
 we predict the value of each node by observing on rest $m-1$ nodes in the structure using exact inferencing implemented in ~\cite{yadav2015business}. We compute the normalized root mean squared error (NRMSE) of the predicted values with respect to the actual value in the data. Averaged out NRMSE, denoted by \textit{$NRMSE_{av}$}, across all nodes is used to measure the predictive accuracy of the learned causal network.

\subsection{Counterfactual accuracy} As discussed in Section~\ref{sec:intro}, counterfactual inference is an important application of causality in many practical domains. We use the counterfactual inference accuracy as a measure to compare different causal graphs learned using various causal discovery algorithms. In order to answer a counterfactual query over a causal graph, we use the approach~\cite{bottou2012counterfactual} of importance sampling to estimate a counterfactual expectation. This approach can be explained briefly via a simple model of an online advertising scenario, as follows:

Consider a Markov factorization of a causal model with four variables given by:
\begin{equation}
 P(w) = P(A)P(C)P(B|A,C)P(D|B)
\end{equation}
where $w$ is shorthand for all variables. Let $D$ be defined as the number of clicks, which is our variable of interest, i.e., target node and $B$ be the ad strategy node. We would like to estimate what the expected click yield $E_D^{'}$ would have been if we had used a different ad strategy. This intervention amounts to replacing the actual factor $P(B|A,C)$ by the counterfactual factor $P^{'}(B|A,C)$ in Markov factorization.
\begin{equation}
P^{'}(w) = P(A)P(C)P^{'}(B|A,C)P(D|B)
\end{equation}
Assuming that the actual factor $P(B|A,C)$ is non-zero everywhere, we can then estimate the counterfactual expected click yield $E_D^{'}$ using the transformation

\begin{equation}
\begin{split}
E_D^{'} = \int_w{D*P^{'}(w)} &= \int_w{D*\frac{P^{'}(B|A,C)}{P(B|A,C)}*P(w)}
\end{split}
\end{equation}

Notice that this can be viewed as the expectation under $P(w)$ after re-weighting the variable of interest $D$ by $\frac{P^{'}(B|A,C)}{P(B|A,C)}$.
Since the expectation is under $P(w)$, it can also be estimated from the observed samples that also follow the original distribution $P(w)$, so:
\begin{equation}
E_D^{'} \propto  \sum_{i=1}^{n} {D_i * \frac{P^{'}(B_i |A_i ,C_i)}{P(B_i|A_i,C_i)}}
\end{equation} 

In principle, we can compute the sample mean of above by normalizing by the sample size; in practice it is observed this can result in bias due to some of the ratios becoming very small, i.e., where $P^{'}(w_i)$ is very small. This is natural since observations under $P(w)$ may be
much more unlikely if evaluated under the modified distribution $P^{'}(w)$. Instability can also result due to rare observations where $P(w)$ is very small.
Therefore in practice one just normalizes by the sum of the re-weighting weights actually used:

\begin{equation}
E_D^{'} \approx  {\sum_{i=1}^{n} {D_i * \frac{P^{'}(B_i |A_i ,C_i)}{P(B_i|A_i,C_i)}} \over \sum_{i=1}^{n} {\frac{P^{'}(B_i |A_i ,C_i)}{P(B_i|A_i,C_i)}}}
\label{counter}
\end{equation}

\section{Benchmarking}\label{sec:experiments}
%\input{Sections/data_description}\label{sec:data}
% !TEX root = ../template.tex
In this section, we present our comparative benchmark results of structure learning algorithms over three datasets having different sample sizes using performance measures for structural, predictive and counterfactual accuracy. Performance evaluations and results for optimal product mix problem is not discussed because of the confidentiality of the data.

\subsection{Data Description} Comprehensive evaluations of causal structure learning algorithms is carried out using two real datasets and one simulated dataset.

\textbf{DREAM4 dataset:} Data from \textit{in-silico} gene regulatory network~\cite{greenfield2010dream4} is used for benchmarking structure learning algorithms. DREAM4 dataset is generated from five networks each, for 10 and 100 nodes respectively with different structures which include feedback loops.
For $10-$node DREAM4 networks, 1 wild type and 10 multi-factorial perturbation data samples are considered as observational dataset while intervention dataset consists of 20 intervention samples, consisting of one knockout and one knockdown per gene. $100-$node DREAM4 networks consist of 201 samples containing 1 wildtype, 100 knockdown and 100 knockout instances.

\textbf{Sachs dataset:} Sachs dataset~\cite{sachs2005causal} consists of observational data collected after general perturbation, which relied on simultaneous measurement of single cell expression profiles of 11 pyrophosphate proteins involved in a signaling pathway of human primary T cells. Sachs dataset contains 1756 observation samples.

\textbf{Simulated dataset:} Data is generated randomly from 10-node Gaussian Bayesian network in \cite{cho2016reconstructing}. Each edge weight is uniformly sampled from $(-1,-.25) \cup (.25,1)$. Base level of each node is randomly sampled from uniform distribution of range $(50,500)$ and noise-level for each node is uniformly sampled from $(5,100)$. 10,000 observational data samples are generated from this model. Knock-out intervention data is also simulated from the same model by knocking out each node sequentially to generate 1000 intervention samples per node.

%%\linebreak
%\paragraph{} We use the pcalg-R implementation for estimating the equivalence class of a directed acyclic graph (DAG) from data.
%
%We vary the alpha,significance level from (0.0 to 1.0)in increments of 0.01 for the individual conditional independence tests to find the best causal graph structure.
%
%\subsection{\textbf{Deciding Optimal Significance Level:}}  
%For deciding the optimal level of significance level we calculate Precision, Recall , F-Score and AUC(Area under the receiver operating characteristic curve) for finding the most accurate underlying causal structure obtained. Our metric for evaluation is similar as in the DREAM4 challenge.
%
%\textbf{Definitions } 
%\linebreak
%\textit{True Positive:} Count of the number of edges, taking into account their orientation, that are present both in the original structure and the learned structure
%
%\textit{True Negative:} Count of the number of edges,taking into account their orientation,that are not present in original structure and the learned structure 
%
%\textit{False Positive:} Count of the number of edges, taking into account their orientation,that are  present in learned structure but not in the original structure  
%
%\textit{False Negative:} Count of the number of edges,  taking into account their orientation, that are not present in the learned structure but are present in the original structure.
%\textit{Precision:}
%\textit{Recall:}
%\textit{F-Score:}
%\textit{Area under the ROC:}

\subsection{Implementation Details}

We have broadly categorized the various causal structure learning techniques into two major classes, as described in Section~\ref{sec:estimating_structures},\ref{sec:CylesAndHidden}. From each class, we have chosen the most prevalent algorithms for which the implementation details are discussed below:
\\
\\
\textbf{Acyclicity and no Latent Variables}

\textbf{Using observational data:} 

PC : We use  pcalg-R \cite{kalisch2010pcalg} for implementation of PC algorithm. Fisher Z-Score conditional independence test is used as conditional independence test for determining skeleton of the graph. By varying values of significance level (0 to 1 in steps of 0.01) of conditional independence tests, different CPDAGs are obtained.
%To estimate the equivalence class of DAG from observational data, using PC-algorithm we use pcalg-R \cite{kalisch2010pcalg} implementation. 
 %The conditional independence of pair of edges given conditioning set $S$, assumes the data to be from multivariate normal distribution and uses Fisher Z-Score conditional independence test for determining skeleton of the graph. By varying values of significance level (0 to 1 in steps of 0.01) of conditional independence tests, different CPDAGs are obtained.

GES : Implementation of GES given in pcalg-R is used.
%Using the greedy equivalence search observational essential graph  \cite{chickering2002optimal} representing the Markov equivalence class of a DAG is estimated, implemented in pcalg-R. 
 $l_0$-penalized Gaussian maximum likelihood estimator is used for scoring the candidate graph networks. 

MMHC :
We use \cite{tsamardinos2006max} implementation of the MMHC Algorithm given in the bnlearn-R package \cite{scutari2012bnlearn}. The BIC criterion is employed as the scoring algorithm and we vary  the target nominal type-I error rate of the conditional independence test for finding the best possible structure CPDAGs.%based on F-Score as the measure for correctness.

GBN: 
We utilize the implementation of GBN given in \cite{cho2016reconstructing}, where the conditional probability distribution (CPD) of each node given the parents is defined to be a linear Gaussian distribution. Output of the algorithm is the graph obtained by model averaging of the set of sampled graph structures drawn from the posterior distribution.
%, where posterior represents how strong is each candidate graph structure to the underlying model given the data. 
 Edge weights of the averaged, weighted resultant graph structure are normalized between 0 to 1, and thresholded at different levels to obtain different CPDAGs corresponding to each threshold level.

\textbf{Using mix of observation and intervention data}

GIES : 
We use \cite{hauser2012characterization} implementation given in pcalg-R for estimating Markov equivalence of a DAG using GIES algorithm. $l$0-penalized Gaussian maximum likelihood estimator is used for scoring candidate causal graph structures.% learned jointly from intervention and observation data.

GBN : We use same implementation as discussed above for this case. %except that both observational and intervention data is used in this case.

\textbf{Active Learning} 

We explore the work of \cite{cho2016reconstructing} which uses active learning to prioritize the interventions for optimal results using GBN framework. It learns the structure using top $n$ interventions out of all given interventions. We experimented on \textit{DREAM4$\_$10} dataset by setting the value of $n$ from 1 to 10, and then choosing the graph with best F-score.
\\
\\
\textbf{Allowing cycles and Latent Variables}
%\subsubsection{Allowing cycles and Latent Variables}

\textbf{Assuming hidden variables:} We use FCI, FCI+ and RFCI algorithms implemented in pcalg-R for comparison across different datasets. For all three algorithms, we vary significance level from 0 to 1 in steps of 0.01 of conditional independence tests to obtain 100 CPDAGs for each algorithm and we choose the graph having best F-score measure.

\textbf{Assuming cycles:} We use CCD algorithm implemented in TETRAD \cite{spirtes2004tetrad} for learning cyclic causal structures. Causal structures corresponding to significance values varying between 0 to 1 with step-size of 0.1 are obtained and the graphs with best F-score is chosen.

\subsection{Experimental Results}

\subsubsection{Structural Accuracy}
Table~\ref{DREAM4-10} shows the structural measure such as F-Score, AUC, SHD, and SID computed on causal graphs learned using various causal structure learning algorithms on real datasets (Dream4 10 nodes and Sachs). It also shows average calculated across all small-sized datasets and across all datasets (small-sized and medium-sized). It is seen that active learning algorithm outperforms all other algorithms for Dream4 datasets (small) in terms of AUC and F-score, while MMHC and PC outperform others in terms of SHD and SID. For medium-sized dataset, i.e., Sachs data, FCI+ seems to outperform every other algorithm in terms of all four structural measures. Algorithms which utilize intervention data (GIES) significantly perform better than their counterparts (GES) which use only observational data.

Table \ref{DREAM4 100} shows the structural accuracy of networks having large number of nodes, 100 in this case. It shows that GES outperforms other algorithms in terms of AUC and F-score. As evident from Table \ref{DREAM4 100}, structural performance drops drastically in case of networks with higher number of nodes. Time taken for execution of structure learning exceeds 24 hour and inferencing algorithms crash for large networks on i7-4600U CPU $@$ 2.10GHz $\times$ 4 system with 8GB RAM. Hence, current state-of-the-art causal structure learning algorithms do not work well for large node networks, present especially in biological domain.

\begin{table*}[]
\centering
\footnotesize
%\fontsize{10}{8}\selectfont
\begin{tabular}{|c|c|c|c|c|c|c|c|c|c|c|c|c|}
\hline
\multirow{2}{*}{\textbf{Dataset}} & \multirow{2}{*}{\textbf{Score}} & \multicolumn{11}{c|}{\textbf{Algorithms}} \\ \cline{3-13} 
 &  & PC & GES & MMHC & GBN & GIES & GBN\_int & FCI & FCI+ & RFCI & CCD & A.L. \\ \hline
\multirow{4}{*}{\textit{DREAM4\_10\_1}} & \textit{F-Score} & 0.41 & 0.29 & 0.36 & 0.28 & 0.36 & 0.38 & 0.41 & 0.41 & 0.41 & 0.35 & \textbf{0.49} \\ \cline{2-13} 
 & \textit{AUC} & 0.55 & 0.59 & 0.54 & 0.56 & 0.64 & 0.67 & 0.27 & 0.34 & 0.34 & 0.53 & \textbf{0.71} \\ \cline{2-13} 
 & \textit{SHD} & 15 & 18 & \textbf{11} & 29 & 17 & 22 & \textbf{11} & 14 & 16 & 13 & 28 \\ \cline{2-13} 
 & \textit{SID} & 38 & 42 & 35 & 43 & 42 & 70 & \textbf{31} & 43 & 45 & 45 & 74 \\ \hline
\multirow{4}{*}{\textit{DREAM4\_10\_2}} & \textit{F-Score} & \textbf{0.45} & 0.28 & 0.34 & 0.37 & 0.29 & 0.44 & 0.42 & 0.44 & 0.44 & 0.42 & \textbf{0.45} \\ \cline{2-13} 
 & \textit{AUC} & 0.57 & 0.56 & 0.57 & 0.70 & 0.57 & 0.73 & 0.18 & 0.30 & 0.30 & 0.57 & \textbf{0.73} \\ \cline{2-13} 
 & \textit{SHD} & 19 & 25 & \textbf{17} & 29 & 21 & 31 & 18 & 18 & 20 & \textbf{17} & 21 \\ \cline{2-13} 
 & \textit{SID} & 65 & 56 & 57 & 44 & 51 & 60 & 69 & 47 & 69 & \textbf{40} & 66 \\ \hline
\multirow{4}{*}{\textit{DREAM4\_10\_3}} & \textit{F-Score} & 0.43 & 0.12 & 0.40 & 0.43 & 0.18 & 0.42 & 0.44 & 0.44 & 0.44 & 0.17 & \textbf{0.47} \\ \cline{2-13} 
 & \textit{AUC} & 0.54 & 0.46 & 0.61 & 0.68 & 0.50 & 0.67 & 0.27 & 0.33 & 0.33 & 0.47 & \textbf{0.74} \\ \cline{2-13} 
 & \textit{SHD} & 14 & 24 & \textbf{11} & 24 & 24 & 21 & 16 & 18 & 17 & 14 & 24 \\ \cline{2-13} 
 & \textit{SID} & 59 & 66 & \textbf{54} & 62 & 67 & 61 & 77 & 79 & 68 & 63 & 76 \\ \hline
\multirow{4}{*}{\textit{DREAM4\_10\_4}} & \textit{F-Score} & 0.32 & 0.12 & 0.27 & 0.34 & 0.29 & 0.75 & 0.30 & 0.32 & 0.32 & 0.44 & \textbf{0.62} \\ \cline{2-13} 
 & \textit{AUC} & 0.49 & 0.47 & 0.55 & 0.66 & 0.60 & 0.80 & 0.25 & 0.33 & 0.33 & 0.50 & \textbf{0.85} \\ \cline{2-13} 
 & \textit{SHD} & 22 & 26 & \textbf{14} & 30 & 25 & 22 & 17 & 24 & 18 & 25 & 26 \\ \cline{2-13} 
 & \textit{SID} & 75 & 59 & 63 & 71 & 59 & 65 & 70 & 56 & 69 & \textbf{29} & 72 \\ \hline
\multirow{4}{*}{\textit{DREAM4\_10\_5}} & \textit{F-Score} & 0.44 & 0.27 & 0.33 & 0.38 & 0.35 & 0.47 & 0.46 & 0.44 & 0.44 & 0.45 & \textbf{0.50} \\ \cline{2-13} 
 & \textit{AUC} & 0.65 & 0.60 & 0.57 & 0.73 & 0.66 & 0.83 & 0.32 & 0.38 & 0.38 & 0.58 & \textbf{0.83} \\ \cline{2-13} 
 & \textit{SHD} & \textbf{9} & 20 & 12 & 20 & 20 & 18 & 11 & 11 & 11 & 10 & 22 \\ \cline{2-13} 
 & \textit{SID} & \textbf{33} & 56 & 42 & 64 & 56 & 70 & 53 & 37 & 38 & 37 & 72 \\ \hline
\multicolumn{1}{|l|}{\multirow{4}{*}{\textit{Avg. DREAM4\_10}}} & \multicolumn{1}{l|}{F-Score} & \multicolumn{1}{l|}{0.41} & \multicolumn{1}{l|}{0.22} & \multicolumn{1}{l|}{0.33} & \multicolumn{1}{l|}{0.36} & \multicolumn{1}{l|}{0.29} & \multicolumn{1}{l|}{0.49} & \multicolumn{1}{l|}{0.40} & \multicolumn{1}{l|}{0.41} & \multicolumn{1}{l|}{0.41} & \multicolumn{1}{l|}{0.36} & \multicolumn{1}{l|}{\textit{\textbf{0.50}}} \\ \cline{2-13} 
\multicolumn{1}{|l|}{} & \multicolumn{1}{l|}{AUC} & \multicolumn{1}{l|}{0.56} & \multicolumn{1}{l|}{0.53} & \multicolumn{1}{l|}{0.57} & \multicolumn{1}{l|}{0.67} & \multicolumn{1}{l|}{0.59} & \multicolumn{1}{l|}{0.73} & \multicolumn{1}{l|}{0.33} & \multicolumn{1}{l|}{0.33} & \multicolumn{1}{l|}{0.33} & \multicolumn{1}{l|}{0.52} & \multicolumn{1}{l|}{\textit{\textbf{0.77}}} \\ \cline{2-13} 
\multicolumn{1}{|l|}{} & \multicolumn{1}{l|}{SHD} & \multicolumn{1}{l|}{15.4} & \multicolumn{1}{l|}{22} & \multicolumn{1}{l|}{\textit{\textbf{13}}} & \multicolumn{1}{l|}{25} & \multicolumn{1}{l|}{20} & \multicolumn{1}{l|}{23} & \multicolumn{1}{l|}{15} & \multicolumn{1}{l|}{17} & \multicolumn{1}{l|}{15} & \multicolumn{1}{l|}{15} & \multicolumn{1}{l|}{24} \\ \cline{2-13} 
\multicolumn{1}{|l|}{} & \multicolumn{1}{l|}{SID} & \multicolumn{1}{l|}{\textit{\textbf{34}}} & \multicolumn{1}{l|}{55} & \multicolumn{1}{l|}{59} & \multicolumn{1}{l|}{74} & \multicolumn{1}{l|}{39} & \multicolumn{1}{l|}{65} & \multicolumn{1}{l|}{59} & \multicolumn{1}{l|}{52} & \multicolumn{1}{l|}{57} & \multicolumn{1}{l|}{42} & \multicolumn{1}{l|}{72} \\ \hline
\multirow{4}{*}{\textit{Sachs dataset}} & \textit{F-Score} & 0.45 & 0.28 & 0.41 & 0.36 & 0.10 & 0.15 & 0.45 & \textbf{0.47} & 0.45 & 0.45 & - \\ \cline{2-13} 
 & \textit{AUC} & 0.67 & 0.57 & 0.63 & 0.61 & 0.40 & 0.46 & \textbf{0.68} & 0.67 & 0.67 & 0.49 & - \\ \cline{2-13} 
 & \textit{SHD} & 19 & 25 & 14 & 16 & 18 & 28 & 22 & \textbf{12} & 18 & 19 & - \\ \cline{2-13} 
 & \textit{SID} & 98 & 85 & 77 & 87 & 78 & 82 & 88 & \textbf{71} & 79 & 108 & - \\ \hline
\multirow{4}{*}{\textit{AVERAGE}} & \textit{F-Score} & 0.42 & 0.23 & 0.35 & 0.36 & 0.26 & 0.44 & 0.41 & 0.42 & 0.42 & 0.38 & \textit{\textbf{0.50}} \\ \cline{2-13} 
 & \textit{AUC} & 0.58 & 0.54 & 0.58 & 0.66 & 0.56 & 0.69 & 0.39 & 0.39 & 0.39 & 0.52 & \textit{\textbf{0.77}} \\ \cline{2-13} 
 & \textit{SHD} & 16 & 23 & \textit{\textbf{13}} & 24 & 20 & 24 & 16 & 16 & 16 & 16 & 24 \\ \cline{2-13} 
 & \textit{SID} & 45 & 60 & 62 & 62 & \textit{\textbf{46}} & 68 & 64 & 55 & 61 & 53 & 72 \\ \hline
\end{tabular}
\caption{Table showing Structural F-Score and AUC for \textit{DREAM4$\_$10} and \textit{Sachs} datasets}
\label{DREAM4-10}
\end{table*}

%\vspace{-10pt}

\begin{table*}[]
\centering
\begin{tabular}{|c|c|c|c|c|c|c|c|}
\hline
\multirow{3}{*}{\textbf{Dataset}} & \multirow{3}{*}{\textbf{SCORE}} & \multicolumn{6}{c|}{\textbf{Algorithms}} \\ \cline{3-8} 
 &  & \multicolumn{2}{c|}{On Observational Data} & On Intervention Data & \multicolumn{3}{c|}{For Hidden Variables} \\ \cline{3-8} 
 &  & PC & GES & GIES & FCI & FCI+ & RFCI \\ \hline
\multirow{2}{*}{\textit{DREAM4\_100\_1}} & \textit{F-Score} & 0.23 & \textbf{0.61} & 0.14 & 0.15 & 0.15 & 0.15 \\ \cline{2-8} 
 & \textit{AUC} & \textbf{0.58} & 0.57 & \textbf{0.58} & 0.18 & 0.31 & 0.31 \\ \hline
\multirow{2}{*}{\textit{DREAM4\_100\_2}} & \textit{F-Score} & 0.15 & \textbf{0.26} & 0.21 & 0.11 & 0.11 & 0.11 \\ \cline{2-8} 
 & \textit{AUC} & 0.008 & \textbf{0.58} & 0.55 & 0.002 & 0.002 & 0.002 \\ \hline
\multirow{2}{*}{\textit{DREAM4\_100\_3}} & \textit{F-Score} & 0.18 & 0.09 & \textbf{0.21} & 0.08 & 0.086 & 0.08 \\ \cline{2-8} 
 & \textit{AUC} & 0.01 & 0.49 & \textbf{0.55} & 0.002 & 0.002 & 0.002 \\ \hline
\multirow{2}{*}{\textit{DREAM4\_100\_4}} & \textit{F-Score} & \textbf{0.17} & 0.09 & 0.07 & \textbf{0.17} & \textbf{0.174} & \textbf{0.17} \\ \cline{2-8} 
 & \textit{AUC} & 0.02 & \textbf{0.48} & 0.45 & 0.006 & 0.004 & 0.004 \\ \hline
\multirow{2}{*}{\textit{DREAM4\_100\_5}} & \textit{F-Score} & \textbf{0.5} & 0.32 & 0.27 & \textbf{0.5} & \textbf{0.5} & \textbf{0.5} \\ \cline{2-8} 
 & \textit{AUC} & 0.05 & \textbf{0.63} & 0.60 & 0.030 & 0.03 & 0.03 \\ \hline
\end{tabular}
\caption{Table showing structural F-score and AUC for \textit{DREAM4$\_$100 datasets}}
\label{DREAM4 100}
\end{table*}

\subsubsection{Predictive Accuracy}
Table \ref{NRMSE} shows the averaged out NRMSE ($NRMSE_{av}$) across all nodes of each learned network. Last column of table \ref{NRMSE} shows the average of $NRMSE_{av}$ across all datasets. Lesser the $NRMSE_{av}$, better is the predictive accuracy. 

In terms of predictive accuracy, GES outperforms all other algorithms for small dataset and does equally good as PC for medium size dataset. Overall $NRMSE_{av}$  of GES for all six dataset is 8.845, followed by GBN learned on observation data and structure learning algorithms taking hidden nodes into account.

From Table \ref{DREAM4-10} and \ref{NRMSE}, it is seen that algorithms having high structural AUC and F-score do not necessary have high inferencing capability. For example, active learning algorithm have highest F-score and AUC for small dataset, but GES outperforms it in terms of predictive accuracy. Similarly, MMHC and PC do better than every algorithm in terms of SHD and SID for small dataset  but their predictive accuracies are low. This emphasizes that rather than just looking at traditional measures of finding structurally best causal network, inferencing capability of the learned causal structure should also be taken into account. Moreover, when true causal structure is not known, predictive accuracy can be used as an effective measure for finding the best causal structure. It is seen that FCI+ which works best for medium size data (Sachs) in terms of structure accuracy, also works reasonably well in terms of predictive accuracy.

\begin{table*}[]
\centering
\scriptsize
\begin{tabular}{|c|c|c|c|c|c|c|c|}
\hline
\textit{\textbf{Algorithm}} & \textit{\textbf{DREAM4\_10\_1}} & \textit{\textbf{DREAM4\_10\_2}} & \textit{\textbf{DREAM4\_10\_3}} & \textit{\textbf{DREAM4\_10\_4}} & \textit{\textbf{DREAM4\_10\_5}} & \multicolumn{1}{l|}{\textit{\textbf{\begin{tabular}[c]{@{}l@{}}Sachs\\ dataset\end{tabular}}}} & \textbf{AVERAGE} \\ \hline
\textit{PC} & 9.01 & 31.43 & 32.79 & 32.13 & 33.15 & 6.2 & 24.1183 \\ \hline
\textit{GES} & 7.62 & 8.3 & 10.18 & 10.43 & 10.24 & 6.3 & 8.845 \\ \hline
\textit{GBN(obs)} & ** & 8.32 & 11.57 & 9.91 & 11.17 & 6.836 & 9.56 \\ \hline
\textit{MMHC} & 16.84 & 10.12 & 22.39 & 16.87 & 12.58 & 8.78 & 14.59 \\ \hline
\textit{GIES} & 10.45 & 11.3 & 12.45 & 11.67 & 10.24 & 7.32 & 10.5716 \\ \hline
\textit{GBN(obs+int)} & 17.54 & 7.5 & 17.5 & 15.18 & 23.52 & 6.26 & 14.58 \\ \hline
\textit{FCI} & 15.72 & 10.8 & 10.04 & 13.08 & 14.66 & 6.59 & 11.815 \\ \hline
\textit{FCIP} & 10.53 & 11.05 & 10.09 & 7.82 & 18.63 & 6.24 & 10.7267 \\ \hline
\textit{RFCI} & 15.16 & 7.76 & 8.09 & 7.87 & 16.33 & 6.39 & 10.2667 \\ \hline
\textit{Active Learning} & 12.27 & 17.37 & 15.58 & 11.47 & 11.97 & * & 13.73 \\ \hline
\end{tabular}
\caption{Averaged NRMSE, obs: (observational dataset) and int: (intervention dataset), ** took more than a day for execution (Results are not presented for the same). }
\label{NRMSE}
\end{table*}

% Similarly, for large dataset (as shown in table~\ref{Simulated dataset}), PC outperforms other algorithms both in terms of structure and predictive accuracy and its performance is close to the original graph. True causal graphs for other datasets are cyclic in nature and hence throughout the analysis, we do not consider comparing predictive accuracy of other datasets with their true causal structures.

\begin{figure}[h]
\centering
\includegraphics[width = 70mm]{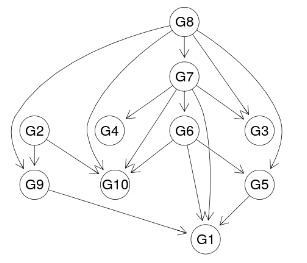}
\caption{Original graph of simulated dataset}
\label{fig:org}
\end{figure}

\begin{figure}[h]
\centering
\includegraphics[width = 90mm]{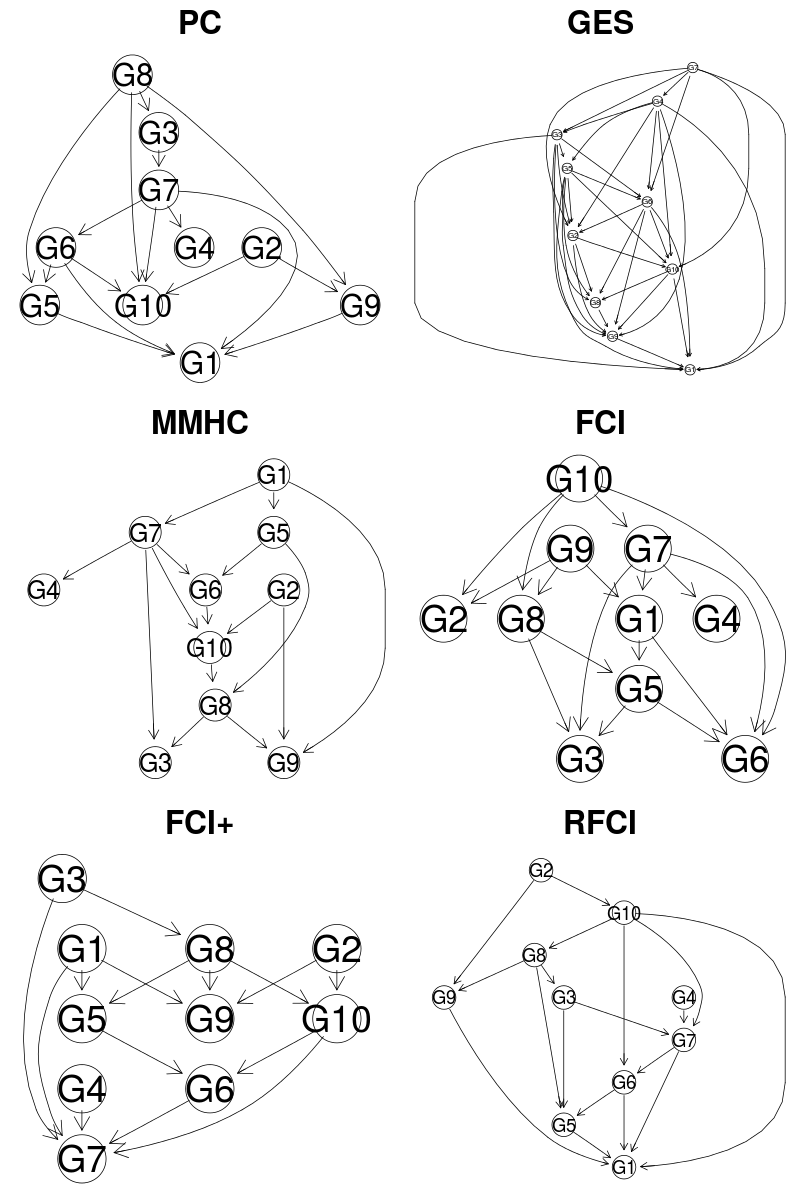}
\caption{Learned graphs using different causal structure learning algorithms}
\label{fig:all}
\end{figure}

\subsubsection{Counterfactual Accuracy}
\begin{figure}[h]
\centering
\includegraphics[width = 90mm]{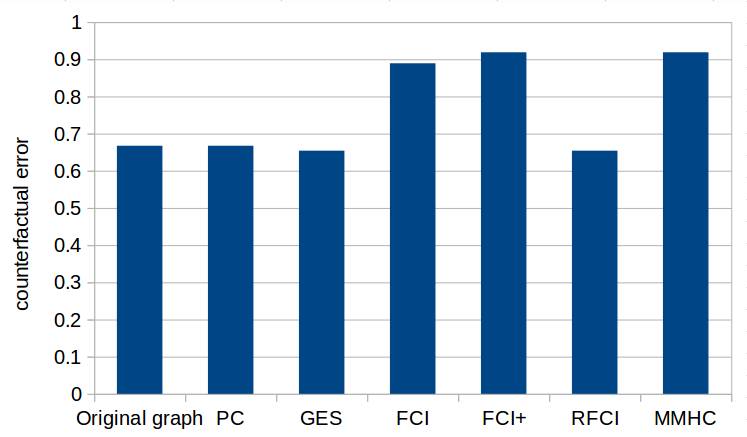}
\caption{Comparison of counterfactual error for graphs learned using various algorithms}
\label{fig:counter_error}
\end{figure}

We compare the counterfactual accuracy over the simulated dataset generated from the network shown in Fig.~\ref{fig:org}. In order to answer the counterfactual question, we perform intervention on node $G6$ by changing incoming edge weight, and sample the data from network. It resulted into updated means of the directly affected nodes $G1, G5,$ and $G10$, called \textit{target nodes}. We calculate the updated means of these three nodes through importance sampling, explained in Section~\ref{sec:metrics}. Figure~\ref{fig:counter_error} shows the average \textit{counterfactual error} (CE) of all three nodes for causal graphs learned using various algorithms (shown in Fig.~\ref{fig:all}). It also shows the CE on original graph. The average counterfactual error is defined as follows:
\begin{equation}
%\begin{split}
  CE =  \frac{\sum_{G_i} (\delta_{act} - \delta_{pred})}{\sum_{G_i} \delta_{act}},  \textrm{    i = 1, 5, and 10}
\end{equation} 
\begin{equation} 
\begin{split}
  & \delta_{act} =  \mid mean_{obs} - mean_{int}\mid\\ 
  & \delta_{pred} =  \mid mean_{obs} - mean_{counter}\mid
\end{split}
\end{equation}
where $mean_{obs}$ is the mean of node before intervention, $mean_{int}$ is the mean after performing intervention on $G_6$ and sample data and $mean_{counter}$ is  mean obtained after importance sampling (as discussed in section~\ref{sec:metrics}). Since the direction of counterfactual change for these nodes is always correct - hence comparing absolute error.

Figure~\ref{fig:counter_error} shows that errors in PC, GES, and RFCI are close to those when the actual graph was used to compute counterfactual expectation (which is, in some sense, the best possible), whereas, errors in FCI, FCI+, MMHC are high as compared to original graph.

%\vspace{-20pt} 

 %This can also lead to the application of causal structure learning in practical fields requiring causal inferencing. 

% \begin{table}[]
% \centering
% \begin{tabular}{|c|c|c|c|c|c|}
% \hline
% \multirow{2}{*}{\textit{\textbf{SCORE}}} & \multicolumn{5}{c|}{\textit{\textbf{Algorithms}}} \\ \cline{2-6} 
%  & PC & \textit{GES} & \textit{FCI} & \textit{FCI+} & \textit{True graph} \\ \hline
% F-score & 0.91 & 0.436 & 0.77 & 0.79 & 1 \\ \hline
% AUC & 0.70 & 0.70 & 0.66 & 0.68 & 1 \\ \hline
% SHD & 2 & 27 & 11 & 12 & 0 \\ \hline
% SID & 12 & 33 & 65 & 71 & 0 \\ \hline
% NRMSE & 7.44 & ** & 9.31 & 7.81 & 7.36 \\ \hline
% \end{tabular}
% \caption{Comparing performance measures on simulated dataset as compared to true structure. (**) representing algorithm took longer than 24 hours for execution}
% \label{Simulated dataset}
% \end{table}

\subsection{Comparison of different size datasets}
We study the performances of causal structure learning algorithms based on the number of data samples in the dataset. \textit{DREAM4$\_$10} dataset is \textit{small-sized} dataset having less than 25 data samples, Sachs dataset is a medium sized dataset with around 1200 observation samples, and \textit{Simulated} dataset is \textit{large-sized} having greater than 10,000 observation samples and more than 1000 intervention. From Table \ref{SizeComparison}, we observe that for small sized dataset active learning based algorithm outperforms rest of the algorithms in terms of AUC and F-score, whereas, in terms of SHD and SID MMHC and PC perform better. For predictive accuracy, GES outperforms rest of the algorithm. For medium sized dataset, FCI+ outperform in all performance metrics of structural accuracy and predictive accuracy. For large dataset, PC outperforms everything else in all performance metrics. Thus, better structure leads to a better predictive accuracy for medium to large datasets but this does not holds for small-sized datasets.

\begin{figure}[h]
\centering
\includegraphics[width = 90mm]{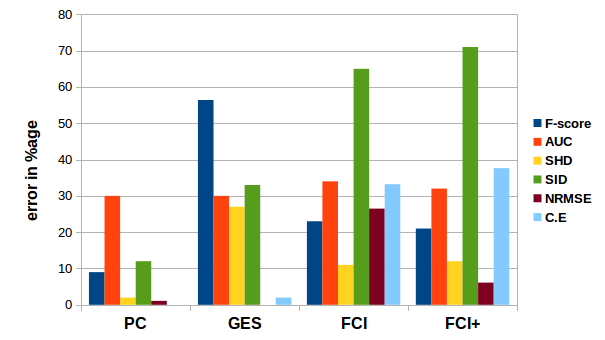}
\caption{Comparison of various algorithms by \% error in performance metrics for simulated dataset. Note- NRMSE for GES algorithm is not present as it is computationally highly inefficient due to highly dense network and took more than 24 hour to complete. }
\label{fig:sim_error}
\end{figure}

For simulated dataset, Figure~\ref{fig:sim_error} shows the comparison of different algorithms using \% error in various perform measures. In order to calculate \% error, we calculate  \%age of absolute difference between the computed value of performance measure and true value to get the percentage error:
\begin{equation}
  \frac{\mid \textrm{computed value} - \textrm{true value} \mid}{\textrm{true value}}*100
\end{equation}
For AUC and F-score, true value is 1 and for SHD and SID, true value is 0. True value in case NRMSE and CE is the predictive accuracy and counterfactual error on true graph, which is 0.76 for NRMSE and 0.667 for CE. Figure~\ref{fig:sim_error} shows that PC outperforms rest of the algorithms in every perform measure.

\begin{table}[]
\centering
\scriptsize
\begin{tabular}{|c|c|c|c|c|c|}
\hline
\textbf{\begin{tabular}[c]{@{}c@{}}Dataset \\ Size\end{tabular}} & \textbf{NRMSE} & \textbf{F-Score} & \textbf{AUC} & \textbf{SHD} & \textbf{SID} \\ \hline
\textit{Small} & \begin{tabular}[c]{@{}c@{}}9.35\\ (GES)\end{tabular} & \begin{tabular}[c]{@{}c@{}}0.50\\ (A.L.)\end{tabular} & \begin{tabular}[c]{@{}c@{}}0.77\\ (A.L.)\end{tabular} & \begin{tabular}[c]{@{}c@{}}13\\ (MMHC)\end{tabular} & \begin{tabular}[c]{@{}c@{}}34\\ (PC)\end{tabular} \\ \hline
\textit{Medium} & \begin{tabular}[c]{@{}c@{}}6.2\\ (PC)\end{tabular} & \begin{tabular}[c]{@{}c@{}}0.47\\ (FCI+)\end{tabular} & \begin{tabular}[c]{@{}c@{}}0.68\\ (FCI)\end{tabular} & \begin{tabular}[c]{@{}c@{}}12\\ (FCI+)\end{tabular} & \begin{tabular}[c]{@{}c@{}}71\\ (FCI+)\end{tabular} \\ \hline
\textit{Large} & \begin{tabular}[c]{@{}c@{}}7.44\\ (PC)\end{tabular} & \begin{tabular}[c]{@{}c@{}}0.91\\ (PC)\end{tabular} & \begin{tabular}[c]{@{}c@{}}0.70\\ (PC,GES)\end{tabular} & \begin{tabular}[c]{@{}c@{}}2\\ (PC)\end{tabular} & \begin{tabular}[c]{@{}c@{}}12\\ (PC)\end{tabular} \\ \hline
%\textbf{AVERAGE} & 7.66 & 0.62 & 0.71 & 8.8 & 40.13 \\ \hline
\end{tabular}
\caption{Best results obtained,algorithm using various performance metrics with respect to size of the datasets. For small datasets, we took an average of measures across datasets}
\label{SizeComparison}
\end{table}

% \usepackage{multirow}
% \begin{table}[]
% \centering
% \begin{tabular}{|c|c|c|c|}
% \hline
% \multirow{2}{*}{\textbf{Algorithm}} & \multicolumn{3}{c|}{\textbf{Target Nodes}} \\ \cline{2-4} 
%  & G1 & G5 & G10 \\ \hline
% \textit{PC} & 63.89 & 97.07 & 108.58 \\ \hline
% \textit{GES} & 60.70 & 57.41 & 88.71 \\ \hline
% \textit{FCI} & 86.56 & 93.37 & 85.66 \\ \hline
% \textit{FCI+} & 85.10 & 102.01 & 98.17 \\ \hline
% \textit{RFCI} & 60.70 & 57.41 & 88.71 \\ \hline
% \textit{MMHC} & 85.10 & 102.01 & 98.17 \\ \hline
% \textit{Original Graph} & 63.89 & 97.07 & 108.58 \\ \hline
% \end{tabular}
% \caption{Change in target value as a percentage of the actual change on intervention(Delta) after performing counterfactual inference on different causal structure learning algorithms}
% \label{Counterfactual}
% \end{table} 

%\section{Discussion}
%\input{Sections/Discussion}
\section{Related Work}\label{sec:rel}
There are number of literature surveys which discuss and compare various causal discovery algorithms theoretically and empirically. A review of constraint based algorithms for causal discovery is presented in \cite{yu2016review}. The review discusses the details of state-of-the-art constraint-based causal structure learning algorithms. Also, it mentions publicly-available software and datasets for constraint-based causal discovery. Various structure learning algorithms for CBNs is discussed in \cite{mahmood2011structure}. This work discusses algorithms of two classes: 1) Using observations and 2) Using mix of observation and interventions. A recent survey paper on causal structure learning~\cite{maathuis2016review} discusses the theory of algorithms on structure learning using observation data, classified into three types: constraint based, score based, and hybrid methods. It also discusses the estimation of causal effects from observational data. A review of recent causal structure learning methods is presented in \cite{kalisch2014causal}, where authors have discussed algorithms mainly from the classes of 1) Using observations data, 2) Using mix of observation and intervention. It also discusses algorithms which assume hidden variable, while it merely touches active learning and algorithms with the assumption of cycles.

% In this paper, we provide an overview of causal discovery techniques, classified into two main categories: 1) as-
% suming acyclicity and no latent variables, and 2) allowing for both cycles and latent variables. In the first category we further divide techniques into three classes, 1) observations only: algorithms which use observation data only to infer causal structure, 2) use mix of observation and intervention: algorithms which use both observations and intervention data, and 3) active learning. To the best of our knoweldge, none of the existing survey/literature compare these categories empirically and theoretically together. We also use two important measures predictive accuracy, and counterfactual accuracy to compare causal discovery algorithms, which are highly useful in applications of causality.

\section{Conclusion and Discussion}\label{sec:conc}
In this paper, we provide an overview of  causal discovery techniques, classified into two main categories: 1) assuming acyclicity and no latent variables, and 2) allowing for both cycles and latent variables. In the first category we further divide techniques into three classes, 1) \textit{observations only}: algorithms which use observation data only to infer causal structure, 2) \textit{use mix of observation and intervention}: algorithms which use both observations and intervention data, and 3) \textit{active learning}. We present a comparative experimental benchmarking of causal discovery techniques using performance metrics constructed from three perspectives: 1) Accuracy of structure learning, 2) Predictive accuracy, and  3) Counterfactual accuracy. Our experiments have been performed on three datasets (two real and one simulated) having different sample sizes; large, medium, and small.

From the comparative benchmarking analysis, we learn that for small-sized datasets, i.e., having lesser number samples, active learning based algorithms outperform the remaining causal structure learning algorithms in terms of AUC and F-score, whereas, MMHC and PC work better in terms of SHD and SID.

For large-sized datasets, only observation based algorithms perform fairly well. It was observed that algorithms which perform poorly in terms structural accuracy on small sized data, still perform well in terms of predictive accuracy. For example, GES algorithm performs poorly in terms of structural measures by learning very complex structures but it performs really well as compared to other algorithms in terms of predictive accuracy. This emphasizes that structural efficiency of algorithm do not guarantee high inferencing capability of the network. However, for medium and large sized datasets, algorithms which outperform in terms of structural accuracy, also perform best for predictive accuracy and counterfactual accuracy. 

Structures learned using mix of observation and intervention data outperform the algorithms using observation data only. For example, GIES algorithm which uses both intervention and observation data outperform GES algorithm which uses only observation data. Also, we have learned that PC outperforms other algorithms in terms of every performance measure if the underlying model of data generation is linear Gaussian and data size is large. Also, none of the existing state of the art causal structure learning algorithms are able to capture the causal relations for the datasets having large number of nodes.

\bibliographystyle{IEEEtran}
\bibliography{scifile} 

\end{document}